\documentclass[10pt,twocolumn,letterpaper]{article}

\usepackage[pagenumbers]{cvpr} % To force page numbers, e.g. for an arXiv version

\usepackage{graphicx}
\usepackage{amsmath}
\usepackage{amssymb}
\usepackage{booktabs}
\usepackage{algorithm}
\usepackage{algorithmic}
\usepackage {makecell}
\usepackage{multirow}
\usepackage{lipsum}
\usepackage[pagebackref,breaklinks,colorlinks]{hyperref}

\usepackage[capitalize]{cleveref}
\crefname{section}{Sec.}{Secs.}
\Crefname{section}{Section}{Sections}
\Crefname{table}{Table}{Tables}
\crefname{table}{Tab.}{Tabs.}

\def\cvprPaperID{5394} % *** Enter the CVPR Paper ID here
\def\confName{CVPR}
\def\confYear{2023}

\newcommand\blfootnote[1]{%
\begingroup
\renewcommand\thefootnote{}\footnote{#1}%
\addtocounter{footnote}{-1}%
\endgroup
}

\begin{document}
\title{MM-Diffusion: Learning Multi-Modal Diffusion Models for \\Joint Audio and Video Generation}

\author{Ludan Ruan\textsuperscript{1}\thanks{},
Yiyang Ma\textsuperscript{2},
Huan Yang\textsuperscript{3$\dagger$},
Huiguo He\textsuperscript{3},\\
Bei Liu\textsuperscript{3},
Jianlong Fu\textsuperscript{3}, 
Nicholas Jing Yuan\textsuperscript{3}, 
Qin Jin\textsuperscript{1},
Baining Guo\textsuperscript{3}\\
\textsuperscript{1}Renmin University of China, \textsuperscript{2}Peking University, \textsuperscript{3}Microsoft Research \\
\tt\small \textsuperscript{1}\{ruanld,qinj\}@ruc.edu.cn, \textsuperscript{2}myy12769@pku.edu.cn, \\ 
\tt\small \textsuperscript{3}\{huayan,v-huiguohe,bei.liu,nicholas.yuan,jianf,bainguo\}@microsoft.com
}

\twocolumn[{%
\renewcommand\twocolumn[1][]{#1}%
\maketitle

\begin{center}
    \centering
    \captionsetup{type=figure}
    \includegraphics[width=\linewidth]{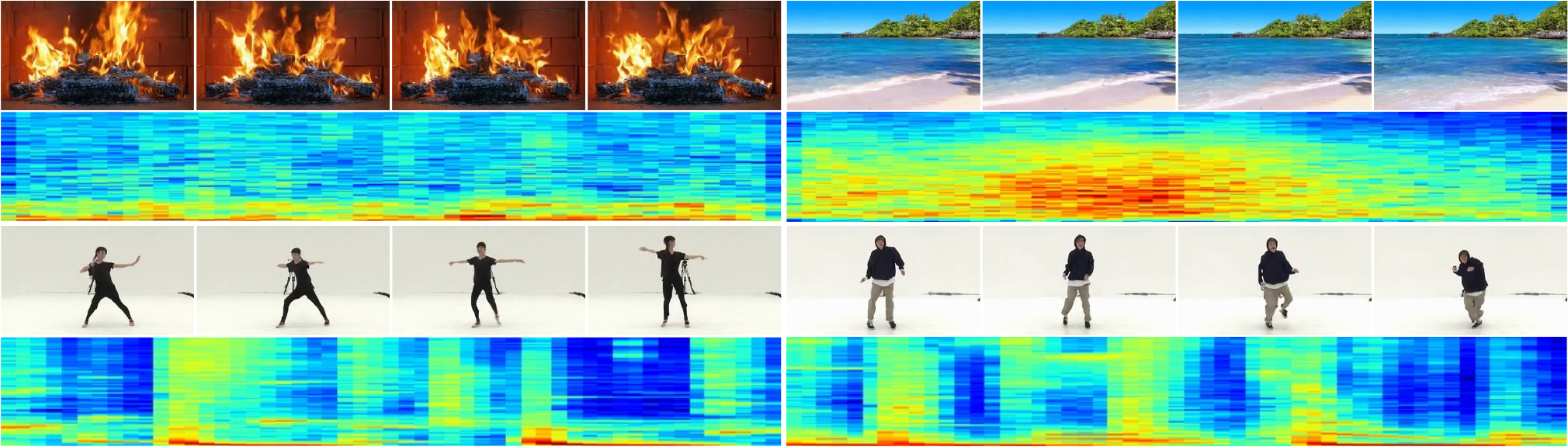}
    \captionof{figure}{Examples of generated video frames ($256 \times 256$) and audio spectrograms from Landscape \cite{lee2022sound} and AIST++ datasets \cite{li2021learn}. We can see vivid  bonfires burning, beautiful sea wave moving, and elegant dancing. Matched audio is generated with video appearances (e.g., the periodical rhythm for dancers). The complete high-fidelity videos and audio can be found in supplementary materials.
    %Add two examples here (video frames, audio wave and anonymous YouTube URL).
    }
\end{center}%
}]

\blfootnote{*This work was performed when Ludan Ruan was visiting Microsoft Research Asia as research interns.} 
\blfootnote{\textsuperscript{$\dagger$}Corressponding author.} 

%%%%%%%%% ABSTRACT
\begin{abstract}
We propose the first joint audio-video generation framework that brings engaging watching and listening experiences simultaneously, towards high-quality realistic videos. To generate joint audio-video pairs, we propose a novel \textbf{M}ulti-\textbf{M}odal \textbf{Diffusion} model (\ie, \textbf{MM-Diffusion}), with two-coupled denoising autoencoders. In contrast to existing single-modal diffusion models, MM-Diffusion consists of a sequential multi-modal U-Net for a joint denoising process by design. Two subnets for audio and video learn to gradually generate aligned audio-video pairs from Gaussian noises. To ensure semantic consistency across modalities, we propose a novel random-shift based attention block bridging over the two subnets, which enables efficient cross-modal alignment, and thus reinforces the audio-video fidelity for each other. Extensive experiments show superior results in unconditional audio-video generation, and zero-shot conditional tasks (e.g., video-to-audio). In particular, we achieve the best FVD and FAD on Landscape and AIST++ dancing datasets. Turing tests of 10k votes further demonstrate dominant preferences for our model.
The code and pre-trained models can be downloaded at \url{https://github.com/researchmm/MM-Diffusion}.

\end{abstract}

%%%%%%%%% BODY TEXT

%-------------------------------------------------------------------------
\section{Introduction}
\label{sec:intro}

AI-powered content generation in image, video, and audio domains has attracted extensive attention in recent years. For example, DALL$\cdot$E 2 \cite{DALLE2} and DiffWave \cite{DiffWave} can create vivid art images and produce high-fidelity audio, respectively. However, such generated content can only provide single-modality experiences either in vision or audition. There are still large gaps with plentiful human-created contents on the Web which often involve multi-modal contents, and can provide engaging experiences for humans to perceive from both sight and hearing. In this paper, we take one natural step forward to study a novel multi-modality generation task, in particular focusing on joint audio-video generation in the open domain.

Recent advances in generative models have been achieved by using diffusion models\cite{OriginalDiffusionModels, DDPM}. From task-level perspectives, these models can be divided into two categories: unconditional and conditional diffusion models. In particular, unconditional diffusion models generate images and videos by taking the noises sampled from Gaussian distributions \cite{DDPM} as input. Conditional models usually import the sampled noises combined with embedding features from one modality, and generate the other modality as outputs, such as text-to-image \cite{DALLE2, Imagen,UMMD}, text-to-video \cite{Make_a_Video, ImagenVideo}, audio-to-video \cite{CDCD}, etc. However, most of the existing diffusion models can only generate single-modality content. How to utilize diffusion models for multi-modality generation still remains rarely explored.

The challenges of designing multi-modal diffusion models mainly lie in the following two aspects. First, video and audio are two distinct modalities with different data patterns. In particular, videos are usually represented by 3D signals indicating RGB values in both spatial (\ie, height $\times$ width) and temporal dimensions, while audio is in 1D waveform digits across the temporal dimension. How to process them in parallel within one joint diffusion model remains a problem. Second, video and audio are synchronous in temporal dimension in real videos, which requires models to be able to capture the relevance between these two modalities and encourage their mutual influence on each other.

To solve the above challenges, we propose the first \textbf{M}ulti-\textbf{M}odal \textbf{Diffusion} model (\ie, \textbf{MM-Diffusion}) consisting of two-coupled denoising autoencoders for joint audio-video generation. Less-noisy samples from each modality (\eg, audio) at time step $t-1$, are generated by implicitly denoising the outputs from both modalities (audio and video) at time step $t$. Such a design enables a joint distribution over both modalities to be learned. To further learn the semantic synchronousness, we propose a novel cross-modal attention block to ensure the generated video frames and audio segments can be correlated at each moment. We design an efficient random-shift mechanism that conducts cross-attention between a given video frame and a randomly-sampled audio segment in a neighboring period, which greatly reduces temporal redundancies in video and audio and facilitates cross-modal interactions efficiently.

To verify the proposed \textbf{MM-Diffusion} model, we conduct extensive experiments on Landscape dataset \cite{lee2022sound}, and AIST++ dancing dataset \cite{li2021learn}. Evaluation results over SOTA modality-specific (video or audio) unconditional generation models show the superiority of our model, with significant visual and audio gains of $25.0\%$ and $32.9\%$ by FVD and FAD, respectively, on Landscape dataset. Superior performances can be also observed in AIST++ dataset\cite{li2021learn}, with large gains of $56.7\%$ and $37.7\%$ by FVD and FAD, respectively, over previous SOTA models. We further demonstrate the capability of zero-shot conditional generation for our model, without any task-driven fine-tuning. Moreover, Turing tests of 10k votes further verify the high-fidelity performance of our results for common users. 

\section{Related Work}
\label{sec:related_work}
\paragraph{Diffusion Probabilistic Models.} Diffusion Probabilistic Models (DPMs) \cite{OriginalDiffusionModels, DDPM} are a new type of generative model which have achieved impressive results. They consist of a forward process (mapping signal to noise) and a reverse process (mapping noise to signal). It has further proved that the forward and reverse processes of DPMs can be done by solving differential equations \cite{DiffSDE}. They usually perform better with a reweighted objective during training \cite{DDPM}. In the aspect of generation quality and diversity, DPMs have outperformed other generative models with appropriate design of the denoising model \cite{ADM}. It has shown that DPMs can perform well in several image generation tasks, such as image inpainting \cite{RePaint}, super-resolution \cite{SR3,SR_Transformer,ST_Video_SR}, image restoration \cite{DDRM}, image-to-image translation \cite{Palette}, \etc. Because of the character of DPMs which infer the denoising model repeatedly hundreds of times, their sampling speed of them is slow compared to other generative models such as GANs \cite{GAN} and VAEs \cite{VAE}. In order to make DPMs more practical, many methods have been proposed. Denoising Diffusion Implicit Models \cite{DDIM} first proposed a method of sampling through a DPM in an implicit way and accelerated the sampling speed. DPM Solver \cite{DPM-Solver, DPM-Solver++} solved the ordinary differential equation of the reverse process of DPMs \cite{DiffSDE}, gave high-order approximated solutions of these equations, and got high-quality results with only about 10-20 evaluations. Stable Diffusion \cite{StableDiffusion} built DPMs on latent spaces so that the number of pixels was decreased. 
Along with the exploration and perfection of DPMs theories, it is becoming more popular to apply diffusion models in multiple domains.

\paragraph{Cross-Modality Generation.} Cross-modal generation such as text-to-visual~\cite{make_a_scene,Make_a_Video,UMMD,AI_Illus,Face_Animation}, text-to-audio~\cite{audiogen}, audio-to-visual~\cite{avgen,cmcgan,TATS}, visual-to-audio~\cite{avgen,visual2sound,cmcgan,CMT,CDCD}, and visual transfer~\cite{Land_Anima,Blind_SR,TTVFI,TTISR,KD_SR,4D_SR,style_trans} has drawn great attention. In the aspect of audio-to-visual generation, Sound2Sight~\cite{Sound2Sight} first proposed a method of generating aligned videos from audio. TATS~\cite{TATS} proposed a time-sensitive transformer projecting audio latent embeddings to video embeddings and achieved SOTA results. For visual-to-audio generation, CMT~\cite{CMT} modeled music rhythms and proposed a method of generating background music corresponding to given videos with controllable music transformers. CDCD~\cite{CDCD} applied DPMs and proposed a contrastive diffusion loss to improve the alignment of the generated audio and the given videos. For bidirectional conditional generation, Chen et al.~\cite{avgen} first propose 2 separate frameworks for audio-to-image and image-to-audio generation. CMCGAN~\cite{cmcgan} further combines audio-image bidirectional transfer with a unified framework and prove it is better than separate frameworks. However, previous works could only generate one modality at a time, while our work can generate two modalities simultaneously.

\section{Approach}
\label{sec:approach}

This section presents our proposed novel \textbf{M}ulti-\textbf{M}odal \textbf{Diffusion} model (i.e., \textbf{MM-Diffusion}) for realistic audio-video joint generation. Before diving into specific designs, we first briefly recap the preliminary knowledge of diffusion models in Sec.~\ref{sec_diffusion_model}. Then, we introduce the proposed \textbf{MM-Diffusion} by further developing vanilla diffusion models to enable semantically-consistent multiple-modality generation in Sec.~\ref{sec_multimodal_diffusion}. After that, we illustrate a coupled U-Net  architecture for joint audio-video data modeling by design in Sec.~\ref{sec_multimodal_unet}. In Sec.~\ref{sec_method_transfer}, we finally discuss the generation capability of our model for conditional multi-modality generation (\ie audio-to-video and video-to-audio) in a zero-shot manner.

\subsection{Preliminaries of Vanilla Diffusion\label{sec_diffusion_model}}
Diffusion-based models~\cite{OriginalDiffusionModels, DDPM} refer to a class of generation algorithms that first transfer a given data distribution $\boldsymbol{x}$ into unstructured noise~(Gaussian noise in practice), and further learn to recover the data distribution by reversing the above forward process.
The original forward process of Denoising Diffusion Probabilistic models~(DDPMs)~\cite{DDPM} is performed over a discrete $T$ time step. Define $x_0$ as a sample from $X$, and $x_T$ as the sample that fits standard Gaussian distribution and is independent from $x_0$ using the Markovian forward process, which can be expressed as follows:
\begin{align}
    q(x_{t}|x_{t-1}) &= \mathcal{N}(x_t; \sqrt{1-\beta_t}x_{t-1}, \beta_t\mathbf{I}), \\
    q(x_{1:T}|x_0) &= \sideset{}{}\prod_{t=1}^T q(x_t|x_{t-1}),\label{q_xt}
    %&= \mathcal{N}(x_t; \sqrt{1- \overline{\alpha}_t}x_0, (1-\overline{\alpha}_t)\mathbf{I}), 
\end{align}
where $t \in [1,T]$, and $\beta_0, \beta_1, ..., \beta_T$ are pre-defined variance schedule sequences. We follow previous works ~\cite{DDPM,DiffSDE} and use the linear noise schedule to increase $\beta_t$.

To recover an original image, learning to reverse the forward process can be simplified as training a model $\theta$ to fit $p_{\theta}(x_{t-1}|x_{t})$ that approximates with $q(x_{t-1}|x_{t}, x_0)$ for all given $t$ and $x_t$. 
Thus the reverse process can be formulated as Equation~\ref{p_theta}, and $x_0$ can be recovered from a probability density $p(x_T)$ with Equation~\ref{p_x0}, which are shown as follows:
 \begin{align}
     p_{\theta}(x_{t-1}|x_t) &= \mathcal{N}(x_{t-1}; \mu_{\theta}(x_t,t), \Sigma_{\theta}(x_t, t)),  \label{p_theta} \\
     p_{\theta}(x_{0:T}) &= p(x_T) \sideset{}{}\prod_{t=1}^T p_{\theta}(x_{t-1}|x_t),\label{p_x0}
\end{align}
where $\mu_{\theta}$ denotes the Gaussian mean value predicted by $\theta$. And finally, $x_0$ can be obtained. In practice, we remove the variance prediction as it only leads to minor improvement~\cite{IDDP, analytic_dpm}. We also omit this term in the following. 

\begin{figure}[t]
  \centering
  \includegraphics[width=\linewidth]{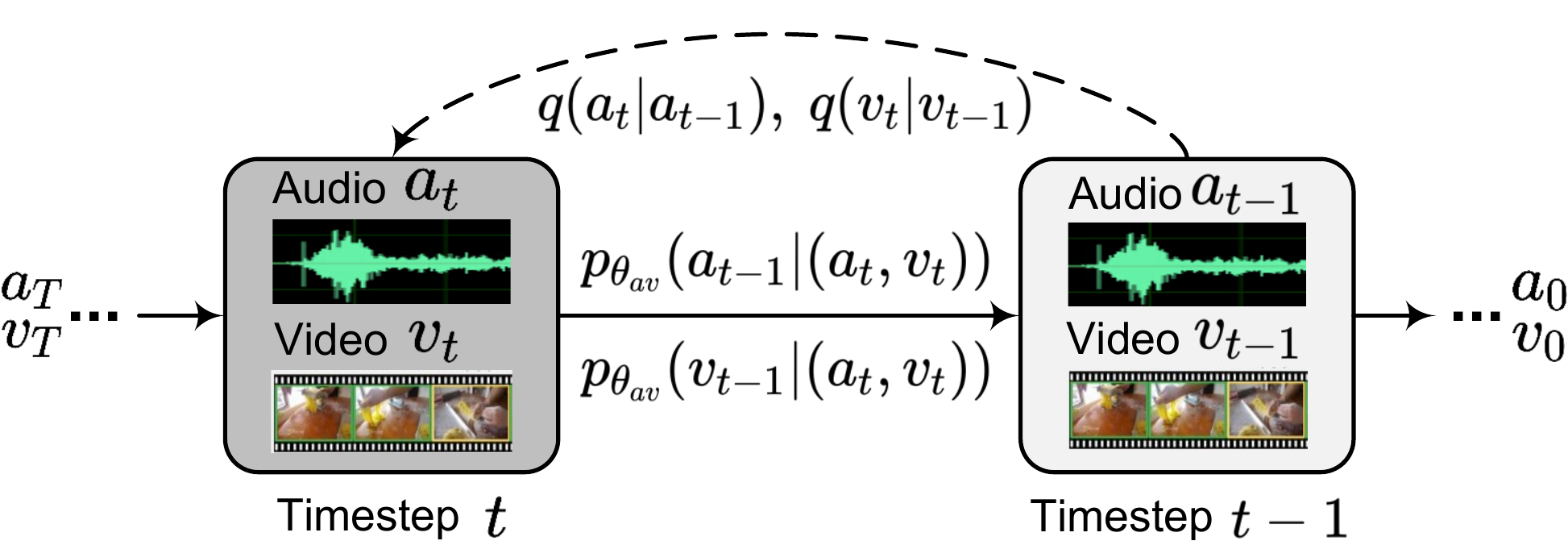}
   
  \caption{An illustration of multi-modal denoising diffusion process.  Forward diffusion (dotted arrow) maps audio \& video data to noise independently, while the reverse process (solid arrow) gradually reconstructs multi-modal contents by a unified model $\theta_{av}$.}
  \label{fig:mm-diffusion}
\end{figure}

\begin{figure*}[t]
  \centering
  \includegraphics[width=\linewidth]{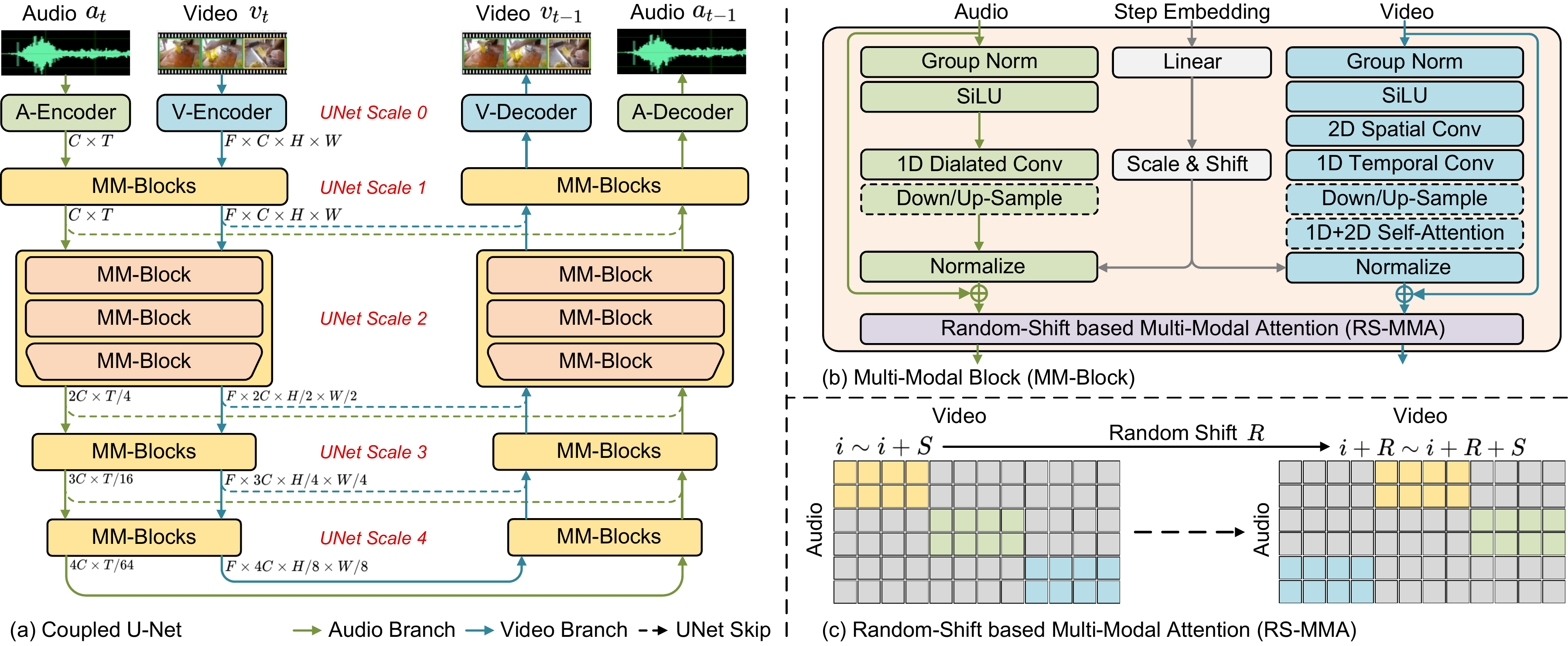}
   
  \caption{Overview of the proposed \textbf{MM-Diffusion} framework. Coupled U-Net contains coupled audio and video streams (indicated by green and blue blocks respectively) at each denoising diffusion step in (a). Each MM-Block encodes audio and video by 1D dilated audio convolutions, and 2D+1D spatial-temporal visual convolutions in (b).  An efficient random-shift based multi-modal attention module is further proposed in (c) to facilitate specific inter-modality alignment and avoid redundant computations.}
  \label{fig:mm-unetn}
\end{figure*}
\subsection{Multi-Modal Diffusion Models} \label{sec_multimodal_diffusion}
With the forward and reverse process in diffusion defined above, we further present the proposed MM-Diffusion formulations in this section. As shown in Figure~\ref{fig:mm-diffusion}, different from the vanilla diffusion where a single modality is generated, our target is to recover two consistent modalities (\ie, audio and video) within one diffusion process. 

Given a paired data $(a,v)$ from a 1D audio set $A$ ($a\in A$), and a 3D video set $V$ ($v\in V$), we consider that the forward processes of each modality are independent, since they are in different distributions. Taking the audio $a$ as an example, its forward process at time step $t$ is defined as:
\begin{align}
    q(a_{t}|a_{t-1}) &= \mathcal{N}_a(a_t; \sqrt{1-\beta_t}a_{t-1}, \beta_t\mathbf{I}), \ \ t \in [1,T].  \label{equ_audio_forward} 
\end{align}
For simplicity, we omit the forward process for videos $v$, as they share a similar formulation. We can further calculate any $a_t, v_t$ using Equation~\ref{q_xt}. 
It is worth noting that we empirically set a shared schedule for hyper-parameters $\beta$ across audio and video to simplify the process definition.

Different from the forward process that models audio and video independently, the correlation between the two modalities should be considered during their reverse processes. Therefore, instead of directly fitting $q(a_{t-1}|a_{t}, a_0)$ and $q(v_{t-1}|v_{t}, v_0)$, we propose a \textbf{unified model} $\theta_{av}$ to take both modalities as inputs and reinforce audio and video generation quality for each other. In particular, for a given time step $t$, the reverse process $p_{\theta_{av}}(a_{t-1}| (a_t, v_t))$ for obtaining $a_{t-1}$ in audio domains  is formulated as follows:
\begin{align}
    p_{\theta_{av}}(a_{t-1}| (a_t, v_t)) = \mathcal{N}(a_{t-1}; \mu_{\theta_{av}}(a_t,v_t,t)),
\end{align} 
where $a_{t-1}$ is generated from a Gaussian distribution jointly determined by both $a_t$ and $v_t$.
To optimize the whole network, we use $\epsilon$-prediction that is defined as:
 \begin{equation}
     \begin{aligned}
         \mathcal{L}_{\theta_{av}} = \mathbb{E}_{\epsilon  \sim \mathcal{N}_a(0, I)}\left[\lambda(t)  \left|\left|\tilde{\epsilon}_{\theta}(a_t, v_t, t) - \epsilon  \right|\right|_2^2\right],
     \end{aligned}
 \end{equation}
where $t \in [0, T]$, and $\lambda_t$ is an optional weighting function. We omit the video formulations since they share similar representations to audio.

The core advantage of multi-modality generation lies in the unified model $\theta_{av}$ that enables jointly reconstructing audio-video pairs from independent Gaussian distributions. Our designed model MM-Diffusion is capable of adapting these two types of input modalities with completely different shapes and patterns. 

\subsection{Coupled U-Net for Joint Audio-Video Denoising}\label{sec_multimodal_unet}
Previous works~\cite{DDPM, ADM, DiffWave, StableDiffusion} have demonstrated the effectiveness of using U-Nets as the model architecture to generate single modality~(\eg, 2D U-Net~\cite{DDPM, ADM} for image generation and 1D U-Net~\cite{DiffWave} for audio generation). Inspired by these works, we propose a coupled U-Net (shown in Figure~\ref{fig:mm-unetn} (a)), which consists of two single-modal U-Nets for audio and video generation. In particular, we formulate input audio and video as a tensor pair $(a, v) \in (A, V)$. On one hand, $a \in \mathbb{R}^{C \times T}$ refers to audio input, where $C$ and $T$ are the channel and temporal dimensions, respectively. On the other, $v \in \mathbb{R}^{ F \times C \times H \times W}$ refers to video input, where $F$, $C$, $H$, and $W$ are frame number, channels, height, and width dimensions, respectively.

\paragraph{Efficient Multi-Modal Blocks.} As shown in Figure~\ref{fig:mm-unetn} (b), for \textbf{video} sub-network design, to efficiently model the spatial and temporal information, we follow Jonathan et al.~\cite{VideoDiffusion} to decompose the spatial and temporal dimensions. 
Specifically, we stack 1D convolutions followed by 2D convolutions as video encoders instead of using the heavy 3D convolutions. Similarly, video attention modules are composed of 2D and 1D attention. Different from videos, the \textbf{audio} signal is a 1D long sequence with higher demand for long-term dependency modeling. Therefore, we have two special designs for audio blocks. First, inspired by Kong~\cite{DiffWave}, we stack dilated convolution layers instead of adopting pure 1D convolutions. The dilation is doubled from $1$ to $2^{N}$, where $N$ is a hyper-parameter. Second, we delete all temporal attentions in audio blocks, which are computationally heavy and showed a limited effect in our preliminary experiments. Previous studies \cite{DiffWave, ImagenVideo} drew similar conclusions as well.

\paragraph{Random-Shift based Multi-Modal Attention.} To bridge the two sub-networks of audio and video, and jointly learn their alignment, the most straightforward way is to perform cross-attention to their features. However, the original attention map for these two modalities is too huge to calculate, with the computational complexity of ${O((F \times H \times W )\times T)}$. Meanwhile, both video and audio are temporal redundant, which means that not all cross-modal attention computations are necessary. 

To solve the above issues, we propose a \textbf{M}ulti-\textbf{M}odal \textbf{A}ttention mechanism with \textbf{R}andom \textbf{S}hift-based attention masks to align video and audio in an efficient way (denoted as \textbf{RS-MMA}), as shown in Figure~\ref{fig:mm-unetn} (c). Specifically, given the $l^{th}$ layer of the coupled U-Nets, with its output of the shape $\{H^l, W^l, C^l, T^l\}$, a 3D video input tensor $v$ with $F$ frames is represented by $F \times H^l \times W^l$ patches, and a 1D audio input tensor is represented by $C^l \times T^l$.

To better align video frames and audio signals, we propose a random-shift attention scheme with following steps:

\noindent \textbf{Step I}: we first split audio stream into segments $\{a_1, a_2, ...a_{F}\}$ along the time steps of video frames, where each segment $a_i$ is in the shape of ${C^l \times \frac{T^l}{F}}$. 

\noindent  \textbf{Step II}: we set a window size $S$ that is much smaller than frame number $F$, and set a random-shift number $R \in [0, F-S]$. 
The attention weights from audio to video are calculated between each audio segment $a_{i}$ and video segment $v_{j}$ starting from frame $f_s$ to frame $f_e$, where  $f_s = (i+R) \% F$ and $f_e=(i+R+S) \% F$. 

\noindent  \textbf{Step III}: the cross attention of audio segment $a_{i}$ and sampled video segment $v_{j}=v_{f_s :f_e}$ is formulated as:
\begin{align}
    {\rm MMA}(a_i,v_j) &= {\rm softmax}(\frac{Q^a_i {K^{vT}_j}}{\sqrt{d_k}}) V^v_j,  \\
    K^v_j &= {\rm linear}({\rm flatten}(v_j)),
\end{align}
where $d_k$ is the dimension of $K$. We omit ${\rm MMA}(v_j, a_i)$, since the cross attention from video to audio is symmetrical. 

This attention mechanism brings two advantages. First, by using such designs, the computation complexity can be reduced to ${O((S \times H \times W) \times ( S \times \frac{T}{F}))}$. Second, the design maintains global attention capabilities within a neighboring period. Since Multi-Modal Diffusion allows iterating from step $T$ to step $0$, video and audio can fully interact with each other during the reverse process. In practice, we set a smaller $S$ at the top of the U-Net to capture fine-grained correspondence, and a larger $S$ at the bottom of the U-Net to capture high-level semantic correspondence in an adaptive way. Detailed settings are described in experiments.

\subsection{Zero-Shot Transfer to Conditional Generation} \label{sec_method_transfer}
Although the MM-Diffusion model is trained for unconditional audio-video pair generation, it can also be utilized for conditional generation (\ie, audio-to-video or video-to-audio) in a zero-shot transfer manner. Because the model has learned the correlation between these two modalities, a strong zero-shot conditional generation performance can help to verify the superior modeling capability of MM-Diffusion. In practice, inspired by Video Diffusion~\cite{VideoDiffusion}, we take two ways for conditional generation, including a replacement-based method and an improved gradient-guided method.

For \textbf{replacement-based} method, to generate an audio $a$ conditioned by a video $v$, \ie, $a \sim p_{\theta_{av}}(a|v)$, we replace $v$ from the reverse process $ p_{\theta_{av}}(a_t|(a_{t+1}, v_{t+1}))$ with samples in forward process $q(\hat{v}_{t+1}|v)$ at each diffusion step $t$. A similar operation can be conducted for video-to-audio generation. However, the replacement-based method predicts the target audio distribution from $a_t \sim \mathbb{E}_{q}(a_t|(a_{t+1}, \hat{v}_{t+1}))$, while the original $v$ that can intuitively provides stronger conditional guidance is neglected. Therefore, we add this condition and reformulate it as \textbf{gradient-guided} method as follows:
\begin{align}
    \mathbb{E}_{q}(a_t|(a_{t+1}, \hat{v}_{t+1}, v)) = \mathbb{E}_{q}(a_t|(a_{t+1}, \hat{v}_{t+1})) + \nonumber \\
     \frac{1}{\sqrt{1-\overline{\alpha}_t}} \nabla_{a_t} \text{log} q(v_t|(a_{t+1}, \hat{v}_{t+1})),
     %\left|\left|a_t - \hat{a_t} \right|\right|_2^2
\end{align}
where $\alpha = 1-\beta$ and $\overline{\alpha_t} = \alpha_t *\alpha_{t-1}*... *\alpha_1$.
Thus, we get generated audio $\tilde{a}_t$ from the following formulation:
\begin{align}
    a_t, v_t & = \theta_{av}(a_{t+1}, \hat{v}_{t+1}), \\
    \tilde{a}_t & = a_t - \lambda \sqrt{1-\overline{\alpha}_t} \nabla_{a_t} \left|\left|v_t - \hat{v}_t  \right|\right|_2^2.
\end{align}
This formulation is also similar to classifier-free conditional generation~\cite{classifier_free_diffusion}, in which $\lambda$ plays the role of a gradient weight to control the intensity of the conditioning. The main difference is that traditional conditional generation models often require explicit training to fit the condition data. Thus their updating process of the sampling procedure does not need to change the condition.
On contrary, to fit the unconditional training process, the conditional input of our gradient-guided method requires constant replacement as the backward process progresses. As a result, we do not need additional training to adapt to conditional inputs which shows a significant merit.

\section{Experiments}
\label{sec:experiments}
In this section, we evaluate the proposed \textbf{MM-Diffusion} model, and compare its joint audio and video generation performance with the SOTA generative models. Visual results can be found in Figure~\ref{fig:gallery}, and more results in the open domain can be found in supplementary materials. 

\subsection{Implementation Details}
\noindent \textbf{Diffusion model.} %As mentioned in Sec. \ref{sec_diffusion_model}, 
To make fair comparison, we follow previous works~\cite{DPM-Solver, DPM-Solver++} to use a linear noise schedule and the noise prediction objective in Sec.~\ref{sec_diffusion_model} for all experiments. The diffusion step $T$ is set as 1,000. 
% Sampling strategy
To accelerate sampling, we use DPM-Solver~\cite{DPM-Solver} as the default sampling method 
unless otherwise specified. 

\noindent \textbf{Model architecture.}
Our whole pipeline contains a coupled U-Net to generate video of $16 \times 3 \times 64 \times 64$, audio of $1 \times 25,600$ , and a \textbf{S}uper \textbf{R}esolution model to scale image from 64 to 256. For the base coupled U-Net, we set 4 scales of MM-Blocks, and each is stacked by 2 normal MM-Blocks and 1 down/up-sample block. 
Only on U-Net scale of [2,3,4], video attention and cross-modal attention are applied, and the window size for cross-modal attention is [1,4,8] corresponding to each scale. 
 
The whole model contains 115.13M parameters. For the SR model, we follow the structure and setting of ADM~\cite{ADM} with 311.03M parameters.
All details of model architecture and training configuration can refer to supplementary materials.

\noindent \textbf{Evaluation.}
To keep consistency, we randomly generate 2,048 samples with each model in the objective evaluation. For fair comparison, metrics are calculated on $64 \times 64$ resolution for all methods. 
In the main results of Sec.~\ref{sec_experiment_result}, we calculate 6 runs on average for reducing randomness. For the ablation study in Sec.~\ref{sec_experiment_ablation}, we sample 2,048 samples from the base coupled U-Net for efficiency. 
% down sample sacle

\subsection{Datasets}
Previous works on video or audio generation mainly focus on one modality. Existing video datasets have problems such as low audio quality, missing audio, and visual-audio mismanagement~(e.g., half audio is missing in UCF-101~\cite{soomro2012ucf101}). 
To facilitate multi-modal generation and extensively compare with different methods, we conduct our experiments on two high-quality video-audio datasets with different types: Landscape~\cite{lee2022sound} and AIST++~\cite{li2021learn}. 

\noindent \textbf{Landscape}
dataset is a high-fidelity audio-video dataset with nature scenes. We crawl 928 source videos from Youtube with the URLs provided by~\cite{lee2022sound}, then divide  into 1,000 non-overlapped clips of 10 seconds. The total duration is about 2.7 hours of 300K frames. Landscape dataset contains 9 diverse scenes, including explosion, fire cracking, raining, splashing water, squishing water, thunder, underwater burbling, waterfall burbling, and wind noise.

\noindent \textbf{AIST++}~\cite{li2021learn} is a subset of AIST dataset~\cite{tsuchida2019aist}, which contains street dance videos with 60 copyright-cleared dancing songs. The dataset includes 1,020 video clips with 5.2 hours duration of about 560K frames in total. To generate clear characters, we uniformly crop out a $1024 \times 1024$ picture from the center of videos for all methods in training.

\subsection{Evaluation Metrics} \label{sec_evaluation_metric}

\noindent \textbf{Objective Evaluation.}
We measure the quality of generated audio and videos separately for the objective evaluation. For videos, we follow prior settings~\cite{yu2022digan,TATS} to use \textbf{Fréchet video distance~(FVD)} and \textbf{kernel video distance~(KVD)} with the I3D~\cite{Joi3dcvpr17} classifier pre-trained on Kinetics-400~\cite{Joi3dcvpr17}.
For audio evaluation, previous works of unconditional audio generation are prone to generate audio in specific domain~(e.g., SC09~\cite{SC09} for spoken digits). Their evaluation metric based on a specifically trained audio classifier is not suitable for our generated audio in the open domain~\cite{CLIP4VLA}. Inspired by FID for image evaluation and FVD for video evaluation, we propose to compute a similar \textbf{Fréchet audio distance (FAD)} between features of generated audio and ground-truth audio~(all FAD numbers need to be multiplied by 1e4). We select AudioCLIP~\cite{Audioclip}, a pre-trained audio model that achieved SOTA in the Environmental Sound Classification  tasks, as the audio feature extractor.

\noindent \textbf{Subjective Evaluation.}
We also conduct user studies on Amazon Mechanical Turk to measure both the quality and relevance of generated audio-video pairs. Specifically, for each audio-video pair, three tasks are formed to measure the quality of the audio, the video, and the relevance of the pair. For each task, we ask users to assign scores ranging from 1 (bad) to 5 (good). We average the scores as the final score, namely \textbf{Mean Opinion Score (MOS)}.
Moreover, we perform Turing Test for audio-video pairs generated by our model and the ground-truth data. We mix them up and ask users to judge whether they are generated or not.

\begin{table}
\centering
\small
\caption{Comparison with single-modal methods on \textbf{Landscape} dataset. * denotes complete ddpm sampling.}
  \begin{tabular}{l|l|lll}
    \toprule
     \# & Method & FVD $\downarrow $& KVD $\downarrow$ & FAD $\downarrow$  \\
      \hline
      1 &Ground-truth &17.83 & -0.12 & 7.51  \\
      \hline
      2 & DIGAN~\cite{yu2022digan} & 305.36 & 19.56 & - \\
      3 & TATS-base~\cite{TATS} &600.30 & 51.54& - \\
      4 & Diffwave~\cite{DiffWave}  & - & - & 14.00 \\
      \hline
      5& Ours-v & 238.33 & 15.14 & - \\
      6& Ours-a & - & - & 13.6 \\
      \hline
      7& Ours& 229.08  &13.26 & \textbf{9.39}  \\  
      8&Ours*&\textbf{117.20}&\textbf{5.78}&10.72\\
    \bottomrule
    
\end{tabular}
\label{tab:comp_withsota_landscape}
\end{table}

\begin{table}
\centering
\small
\caption{Comparison with single-modal generation methods on \textbf{AIST++} dancing dataset.}
  \begin{tabular}{l|l|lll}
    \toprule
     \# & Method & FVD $\downarrow $& KVD $\downarrow$ & FAD $\downarrow$  \\
      \hline
      1 &Ground-truth & 8.73 & 0.036 & 8.46  \\
      \hline
      2 & DIGAN~\cite{yu2022digan} & 119.47 &35.84 &-  \\
      3 & TATS-base~\cite{TATS} &267.24 &41.64 & - \\
      4 & Diffwave~\cite{DiffWave}  &- &- &15.76 \\
      \hline
      5&Ours-v & 184.45 & 33.91 & - \\
      6&Ours-a & - & - & 13.30  \\
      \hline
      7& Ours& 176.55&31.92&12.90\\
      % 8& Ours*& \textbf{51.70} &\textbf{10.22} &\textbf{9.85}  \\
    8& Ours*& \textbf{75.71} &\textbf{11.52} &\textbf{10.69}   \\
    \bottomrule
    
\end{tabular}
\label{tab:comp_withsota_aist}
\end{table}

\begin{figure*}[t]
  \centering
\includegraphics[width=\linewidth]{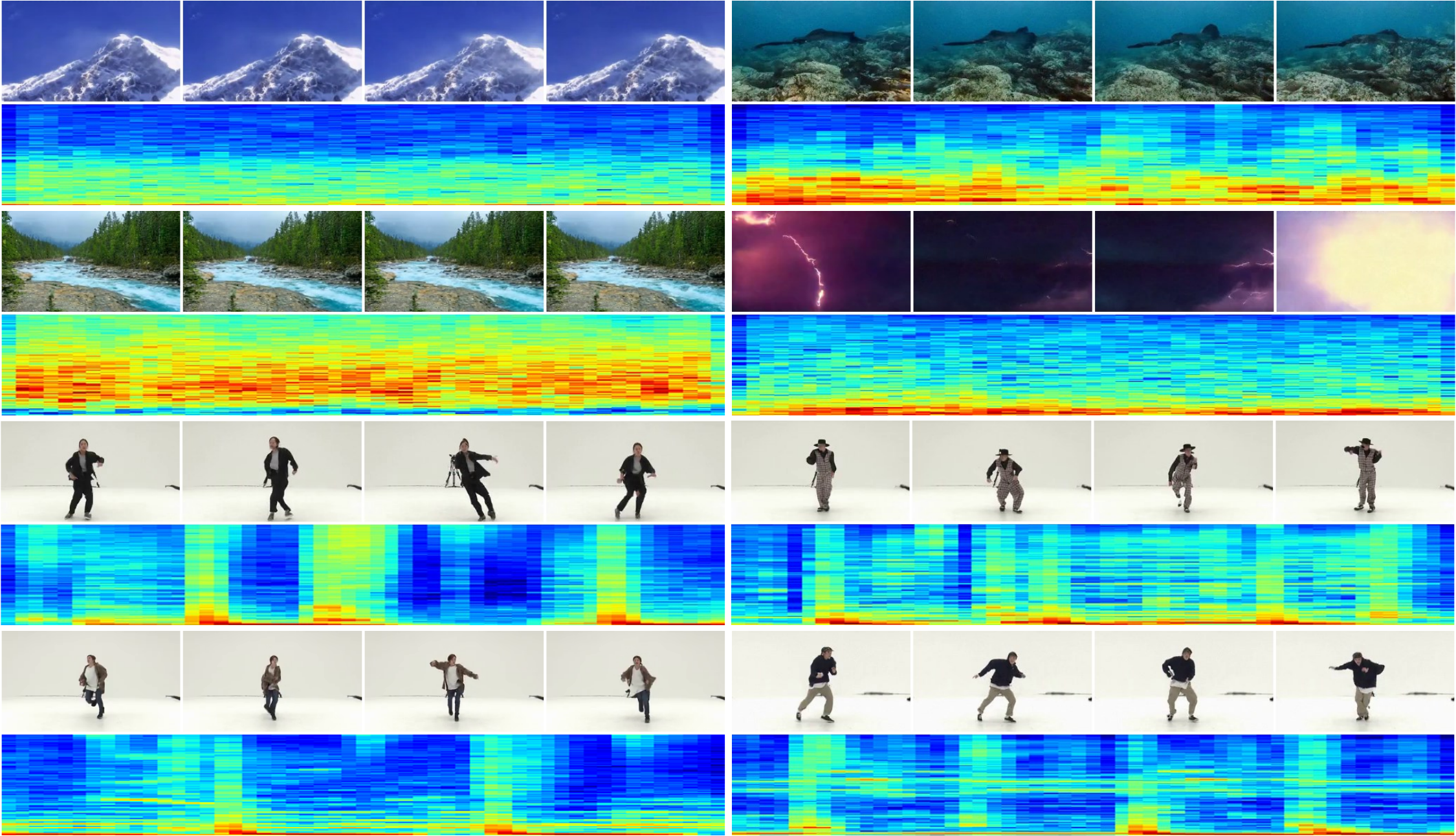}
 
  \caption{More visual examples of generated video frames ($256 \times 256$) with semantic-consistent audio (shown in spectrograms). Some cases vividly show the wind blowing in snow mountains, and some show continuous river sound with beautiful scenes.}
  \label{fig:gallery}
\end{figure*}

\subsection{Objective Comparison with SOTA methods} \label{sec_experiment_result}
To evaluate the quality of audio and video generated by MM-Diffusion, we compare it with SOTA unconditional video generation methods DIGAN~\cite{yu2022digan}, TATS~\cite{TATS}, and audio generation method Diffwave~\cite{DiffWave}. Note that we select these baselines as they are widely-used and have released official codebases for standard replacement on our datasets.
To further explore the effectiveness of joint-learning in MM-Diffusion and to make fair comparisons to single-modality generation with the same backbone, we decompose the coupled U-Nets into audio sub-network~(Ours-a) and video sub-network~(Ours-v) for modality-independent generation. 
The results on Landscape and AIST++ are shown in Table~\ref{tab:comp_withsota_landscape} and Table~\ref{tab:comp_withsota_aist}. 

From these two tables, we can draw the following conclusions: 
%%%%%%TODO
(1) Our model significantly outperforms SOTA single-modal generation methods on both video and audio generation. In particular, our model elevates the SOTA FAD towards ground-truth quality. It demonstrates the effectiveness of the proposed MM-Diffusion and coupled U-Net.
(2) Our model with only video generation (Ours-v) even outperforms SOTA methods DIGAN and TATS-base on most metrics in both tables (comparing \#5 to \#2 and \#3). This indicates that the diffusion-based method can improve the quality of generated videos compared to traditional methods. 
(3) By comparing our full setting (\#7) to one-stream U-Net ($\mathrm{Ours}$-$\mathrm{v}$ and $\mathrm{Ours}$-$\mathrm{a}$), we can see that Coupled U-Nets that jointly learn cross-modality alignment bring further benefit for both video and audio generation. 
Moreover, complete sampling strategy (\#8) will obtain samples of better quality than Dpm-Solver.

\subsection{Ablation Studies} \label{sec_experiment_ablation}
\noindent \textbf{Random-Shift based Multi-modal Attention.}
We have demonstrated the effectiveness of our proposed Random-shift based Multi-Modal Attention mechanism (RS-MMA) in Sec. \ref{sec_experiment_result}. We further conduct two ablation experiments to explore different window sizes and the effectiveness of random shift mechanisms. 
\textbf{(1) Different window sizes.} We first set different window sizes to scale [2,3,4] of the Coupled U-Net. All the experiments are training with 80K steps to save cost and the results are shown in Table~\ref{tab:ablation_multimodal_attention_windowsize}. From the first three lines, we can see that larger window sizes bring more improvement. 
The best performance of the adaptive window size according to the channel scale in U-Net shows the effectiveness of this efficient design, especially for improving video generation quality.
\textbf{(2) Random shift mechanism.}
Table~\ref{tab:rshift} shows the results of whether to use random shift (RS) during training. From the comparison, we can find that RS helps generate audio with better quality compared with no shifting, and the convergence of audio is also accelerated. This also demonstrates that our proposed RS-MMA encourages more efficient joint  cross-modality learning.

Meanwhile, we can see the improvement is more significant in audio quality by using RS. Because the video appearance can provide more information for its paired audio, compared with the effect of audio on the paired video.

\begin{table}[t]
\centering
\small
\caption{Ablation study on the window size of multi-modal attention, corresponding to U-Net scale [2,3,4], at 80k training steps.} 
\vspace{-5px}
  \begin{tabular}{l|c|cccc}
    \toprule
      \# & W-Size&  FVD $\downarrow $& KVD $\downarrow$ & FAD $\downarrow$ & \\
      \hline
      1 & [1,1,1]  & 374.18 & 22.26 & 9.81 \\
      2 & [4,4,4]  & 361.65 & 21.64 & 9.65  \\
      3 & [8,8,8]  & 350.60&21.47 & \textbf{9.50}  \\
      4 & [1,4,8]  & \textbf{303.19} & \textbf{17.26} & 10.20 \\
    \bottomrule
    
\end{tabular}
\label{tab:ablation_multimodal_attention_windowsize}
\end{table}

\begin{figure}
  \centering
  \includegraphics[width=\linewidth]{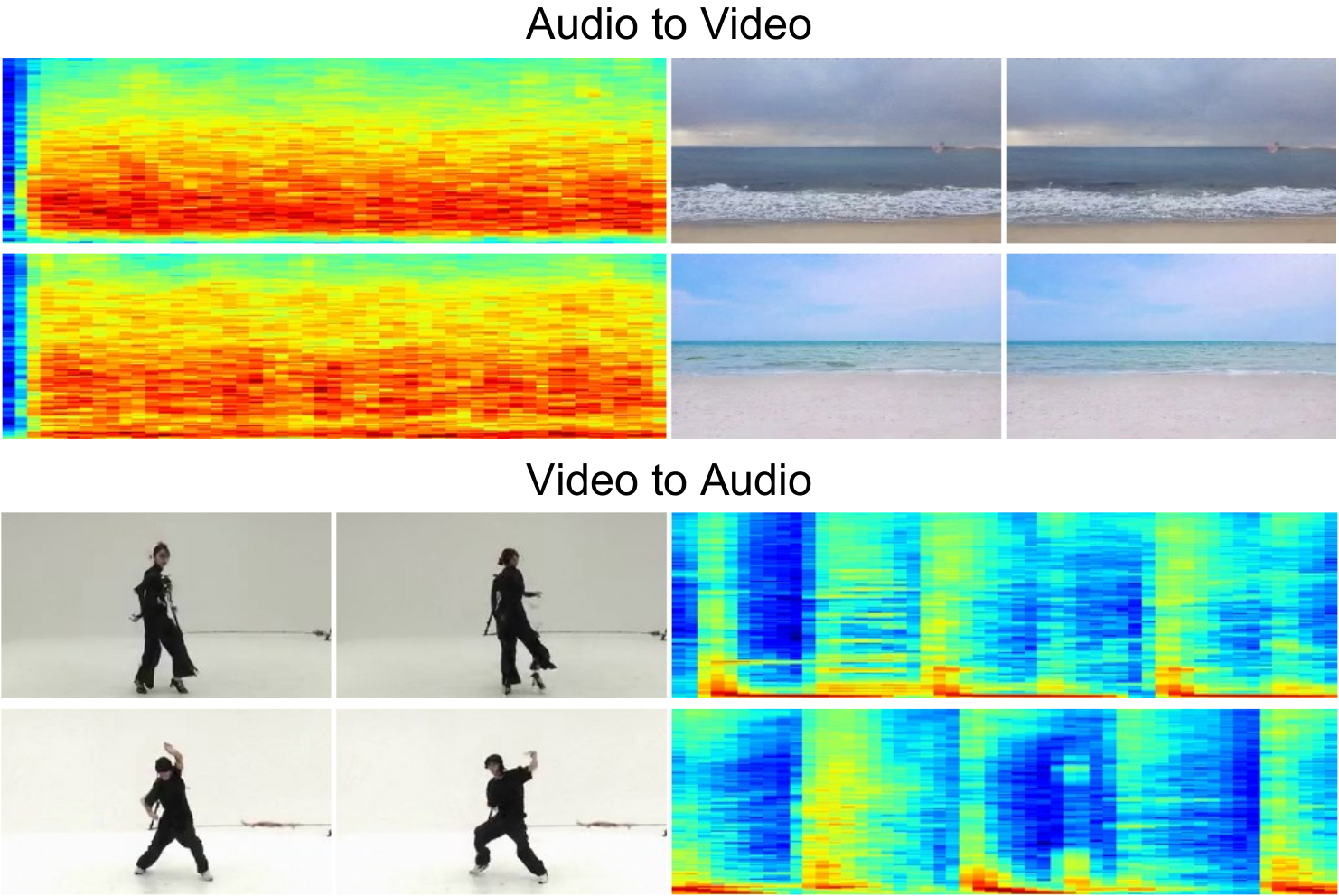}
  \caption{Illustration of several randomly-selected examples generated by zero-shot transferring to conditional generation. We adopt the gradient-guided method for better results.}
  \label{fig:transfer}
\end{figure}

\noindent \textbf{Zero-Shot Conditional Generation.} 
We validate the effectiveness of the two methods for zero-shot transferring and find that both can generate high-quality audio using videos as the condition. 
For audio-based video generation, 

the gradient-guided method is better than the replacement method to get consistent videos that are semantic and temporal aligned with given audio.
The results also show that our model has the ability of modality transfer even without extra training. 
Figure~\ref{fig:transfer} illustrates that our model can generate videos of similar scenes~(sea) from audios with similar patterns, or generate audio that matches the rhythm of the input dancing videos.
This further verifies that our joint learning  can enhance single-modality generation.

\begin{table}[t]
\centering
\small
\caption{Video and audio quality at different training steps, affected by random-shift attention.}
\vspace{-5px}
  \begin{tabular}{l|cc}
    \toprule 
    ~ & \multicolumn{2}{c}{60K/\ 80K/\ 100K Iteration} \\
        \cline{2-3}
       Method & FVD$\downarrow$ & FAD$\downarrow$\\
      \hline
       w/o RS  & 415.50/\ 303.19/\ 271.64  & 12.55/\ 10.20/\ 9.94 \\
       w/ RS  & 440.78/\ 306.42/\ 267.58  & 13.29/\ 9.67/\ 9.10 \\
    \bottomrule
\end{tabular}
\label{tab:rshift}
\end{table}

\subsection{User Studies}
\noindent \textbf{Comparison with other Methods.} As we are the first to jointly generate audio-video pairs, there is no direct baselines to compare. Hence we choose a 2-stage pipeline by using an existing single-modality model. 

In particular, we take a noise-audio-video order as the pipeline. Specifically, we utilize Diffwave \cite{DiffWave} for unconditional audio generation and TATS \cite{TATS} to transfer audio to video. 
For each dataset, we randomly sampled 1,500 audio-video pairs from our model, baseline, and ground-truth data, each with 500 samples. As we explained in Sec. \ref{sec_evaluation_metric}, each pair is divided into 3 tasks. Each task was assigned to 5 users. Thus, we have 45k votes from 9,000 tasks in total. 
From the results in Table \ref{tab:human_study_MOS}, we can see that the quality of audio-video pairs generated by our method on both datasets is much better than the 2-stage baseline method, and our results have much smaller gap with the ground-truth data.
\noindent \textbf{Turing Test.}

To evaluate the realisticness of our generated videos, we further conduct a Turing test. For each dataset, we randomly sampled 500 audio-video pairs from our generated results and ground-truth data, respectively. Each sample was assigned to 5 users and we have 10k votes in total.
From the results shown in Table~\ref{tab:human_study_Turing}, we can see that over 80\% generated sound videos in Landscape can successfully cheat the subjects. Even in AIST++, almost half of the generated sound videos can fool users although the fine-grained parts of persons are difficult to be well generated. This test provides strong validation of the high quality and realism of sound videos we generated for common users.

\begin{table}[t]
\centering
\small
\caption{Mean opinion score (5 is the highest). VQ/AQ denotes video and audio quality. A-V denotes cross-modal alignment.}
\vspace{-5px}
  \begin{tabular}{l|ccc|ccc}
    \toprule 
    ~ &\multicolumn{3}{c}{Landscape} & \multicolumn{3}{c}{AIST++} \\
   \cline{2-4} \cline{5-7} 
       Method & VQ$\uparrow$ & AQ$\uparrow$ & A-V$\uparrow$  & VQ$\uparrow$ & AQ$\uparrow$ & A-V$\uparrow$\\
      \hline
       GT  &3.84 &4.22 &4.52  & 3.79&3.89 &4.15 \\
      2-Stage &1.61 &1.74 &1.72 &2.31 &2.27 &1.81 \\
      Ours& 3.75 &3.93  &4.33 & 3.48&3.50&3.87 \\
    \bottomrule
\end{tabular}
\label{tab:human_study_MOS}
\end{table}

\begin{table}[t]
\centering
\small
\caption{Turing Test on Landscape and AIST++, the number is the percentage of data that is considered to be from real world.}
\vspace{-5px}
  \begin{tabular}{l|cc}
    \toprule 
       ~ & Landscape & AIST++ \\
      \hline
      Ours & 84.9 & 49.6 \\
        Ground-truth & 92.5& 84.7 \\
    \bottomrule
    \end{tabular}
\label{tab:human_study_Turing}
\end{table}

\section{Conclusion}

In this paper, we propose \textbf{MM-Diffusion}, a novel multi-modal diffusion model for joint audio and video generation. Our work pushes the current content generation based on single-modality diffusion models one step forward, and the proposed MM-Diffusion can generate realistic audio and videos in a joint manner. Superior performances are achieved over widely-used audio-video benchmarks by objective evaluations and Turing tests, which can be attributed to the new formulation for multi-modal diffusion, and the designed coupled U-Net. 
In the future, we will add text prompts to guide audio-video generation as a more user-friendly interface, and further develop various video editing techniques (e.g., video inpainting, background music synthesis) by multi-modal diffusion models.

%%%%%%%%% REFERENCES
{\small
\bibliographystyle{ieee_fullname}
\bibliography{egbib}
}
\clearpage
\appendix
\section*{Supplementary Material}
In this supplementary material, we introduce the algorithm details in Sec.~\ref{sec_details}. Next, we propose more details of human study in Sec.~\ref{sec_humanstudy}. Finally, we show more visualization results in Sec.~\ref{sec_visualization}.

\section{Algorithm Details}
\setcounter{table}{0}   %从0开始编号，显示出来表会A1开始编号
\setcounter{figure}{0}
\setcounter{algorithm}{0}
%定义编号格式，在数字序号前加字符“A"
\renewcommand{\thetable}{A.\arabic{table}}
\renewcommand{\thefigure}{A.\arabic{figure}}
\renewcommand{\thealgorithm}{A.\arabic{algorithm}}

\label{sec_details}
In this section, we introduce all the implementation details to ensure the reproducibility of our results. We formally list the details of \textbf{Architecture}, \textbf{Diffusion Process}, \textbf{Training Settings} of the Coupled U-Net and super-resolution network in Table.~\ref{tab:detail}.

\begin{table*}
\centering
\small
 \begin{tabular}{lll}
 \toprule
    & Coupled U-Net & Super-Resolution\\
    \midrule
    \noindent \textbf{Architecture} \\
     Base channel &  128 & 192\\
     Channel scale multiply & 1,2,3,4 & 1,1,2,2,4,4 \\
     Blocks per resolution & 2 + 1down/up sample & 2 + 1down/up sample\\
     Video downsample scale & H/2, W/2 & H/2, W/2\\
     Audio downsample scale & T/4 & N/A\\
     Video attention scale & 2,3,4 & 4,5,6 \\
     % Audio attention scale & None & N/A \\
     Audio conv dilations & 1,2,4,...$2^{10}$  & N/A\\
     Cross-modal attention scale & 2,3,4 & N/A\\
     Cross-modal attention window size & 1,4,8 & N/A \\
     Attention head dimension & 64 & 48\\
     Step embedding dimensions & 128 & 192\\
     Step embedding MLP layers & 2 & 2\\
    
    \midrule
     \textbf{Diffusion Process} \\
     Diffusion noise schedule & linear & linear\\
     Diffusion steps & 1000 & 1000\\
     Prediction target & $\epsilon$ & $\epsilon$\\
     Learn sigma & False & True \\
     Sample method & DPM solver & DDIM \\
     Sample step & N/A & 25 \\
     
    \midrule
     \textbf{Training Settings} \\
     Video sahpe & $16\times 64\times 64 $ & LR: $64\times 64$, HR: $256\times 256$\\
     Video fps & 10 & N/A\\
     Audio shape & $ 1\times 25600 $ & N/A \\
     Audio sample rate & 16,000 Hz & N/A\\
     Augmentation & N/A & Gaussian noise, $\sigma\in[0,20], p=0.5$\\
     & & JPEG compression, $q\in[20, 80], p=0.5$\\
     & & Random flip, $p=0.5$\\
     Weight decay & 0.0 & 0.0\\
     Dropout & 0.1 & 0.1\\  
     Learning rate & 1e-4 & 1e-4 \\
     Batch size & 128 & 48\\
     Training steps & 100,000 & 270,000 \\
     Training hardware & $32\times V100 $ & $8\times V100$\\
     EMA & 0.9999 & 0.9999\\
     \bottomrule
\end{tabular}
\caption{The implementation details of our Coupled U-Net and super-resolution network.}
\label{tab:detail}
\end{table*}

\begin{table*}[t]
\centering
\small
  \begin{tabular}{l|l|l}
    \toprule 
      Score & Video/Audio Quality & Video-Audio Alignment\\
      \hline
      1 & Pure noise, completely unrecognizable content. & \makecell[l]{The video and audio are total noise and \\they are completely irrelevant.} \\
      \hline
      2 & \makecell[l]{The video/audio has development, \\
      but the video/audio type can not be recognized.} & \makecell[l]{The type of video/audio can not be recognised \\and they are irrelevant.} \\
      \hline
      3 & \makecell[l]{Video/audio can be recognized as specific type, \\
      but very unnatural.} & \makecell[l]{The type of video/audio can be recognized \\but they are misaligned.}\\
      \hline
      4 & \makecell[l]{The video/audio is natural, \\but can be recognized as generated content.} &  \makecell[l]{Video and audio are basically matched, \\but the detail in correspondence is lacking.}\\
      \hline
      5 & \makecell[l]{The video/audio is so natural that can not be \\recognized if is from generation or real-world.} & \makecell[l]{The video and audio are consistent \\in detail and very natural.} \\
    \bottomrule
    \end{tabular}
\caption{The score description of MOS for video/audio quality and video-audio alignment.}
\label{tab:turing_MOS}
\end{table*}

\section{Details of Human Study}
\setcounter{table}{0}   %从0开始编号，显示出来表会B1开始编号
\setcounter{figure}{0}
\setcounter{algorithm}{0}
%定义编号格式，在数字序号前加字符“A"
\renewcommand{\thetable}{B.\arabic{table}}
\renewcommand{\thefigure}{B.\arabic{figure}}
\renewcommand{\thealgorithm}{B.\arabic{algorithm}}

\label{sec_humanstudy}
To subjectively evaluate the generative quality of our MM-diffusion, we conduct 2 kinds of human study as written in the main paper: MOS and Turing test.
For MOS, we asked testers to rate the video quality, audio quality and video-audio alignment based on the standards in Table~\ref{tab:turing_MOS}.
For Turing test, we asked the common users to vote the given video: 1). It is generated by machine; 2). It cannot be determined if the video is machine-generated or real; 3). It is real.
We regard the latter two votes as passing the Turing test.

\section{Additional Samples}
\setcounter{table}{0}   %从0开始编号，显示出来表会A1开始编号
\setcounter{figure}{0}
\setcounter{algorithm}{0}
%定义编号格式，在数字序号前加字符“A"
\renewcommand{\thetable}{C.\arabic{table}}
\renewcommand{\thefigure}{C.\arabic{figure}}
\renewcommand{\thealgorithm}{C.\arabic{algorithm}}

\begin{figure*}[t]
  \centering
  \includegraphics[width=\linewidth]{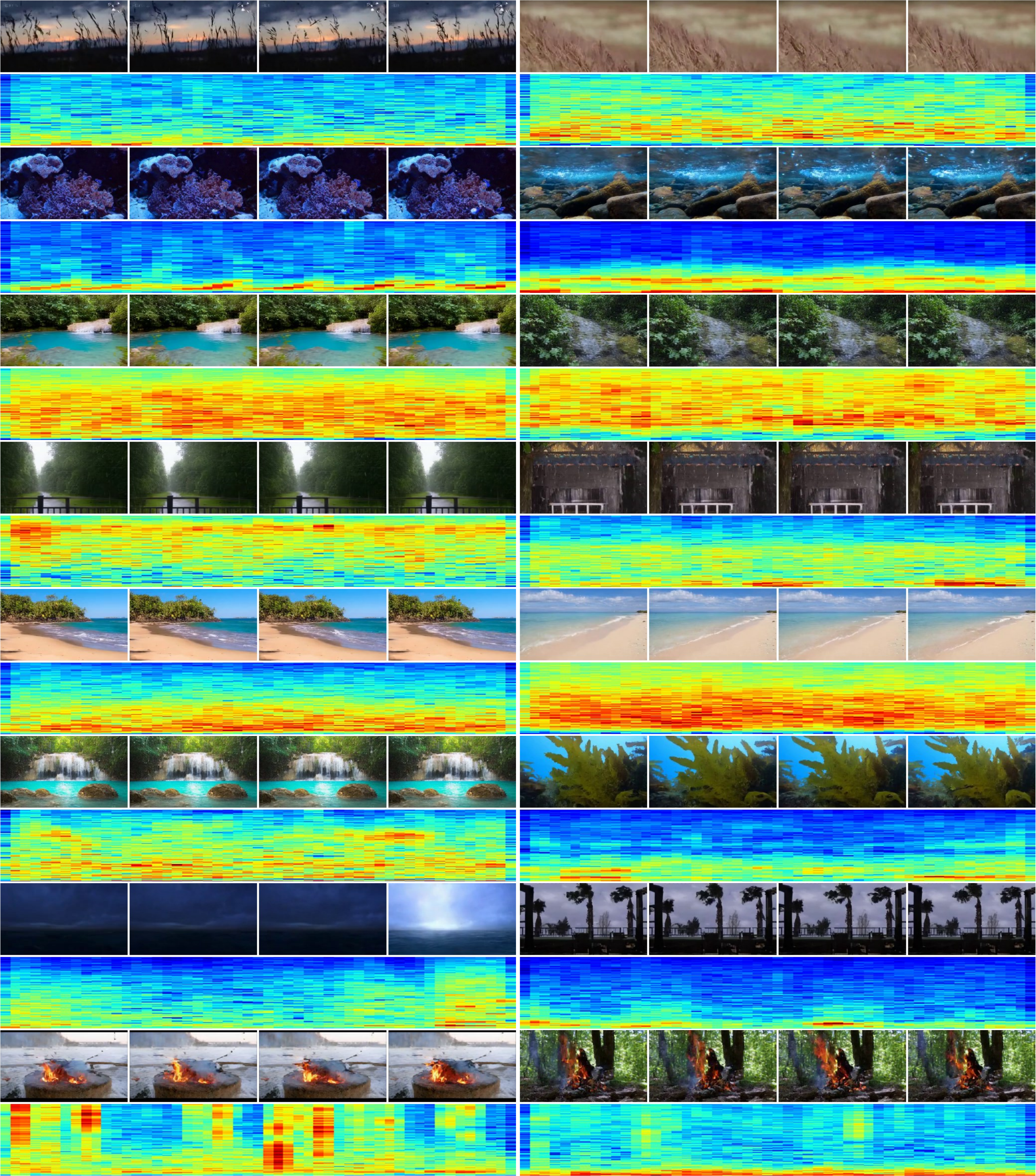}
  \caption{More generation results from Landscape of our MM-Diffusion. The given cases show the scenes of blowing wind, underwater, splashing water, raining, squashing water, waterfall, thunder, and fire cracking respectively.}
  \label{fig:land}
\end{figure*}

\begin{figure*}[t]
  \centering
  \includegraphics[width=\linewidth]{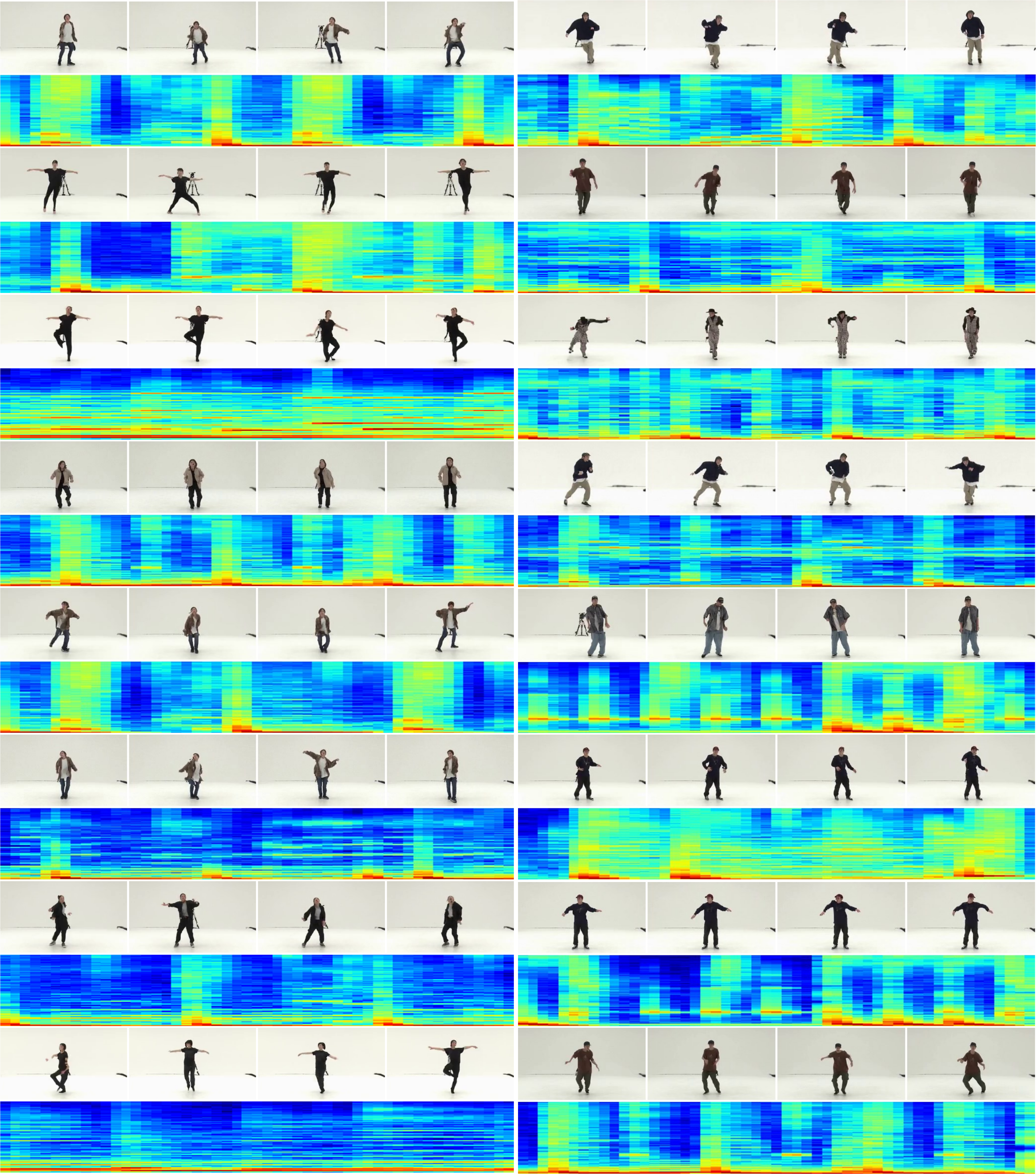}
  \caption{More generation results from AIST++ of our MM-Diffusion. Matched audio is generated with video appearances (e.g., the periodical rhythm for dancers).}
  \label{fig:aist}
\end{figure*}

\label{sec_visualization}
In this section, we show more unconditional generation results of video-audio pairs from Landscape, AIST++ and AudioSet~\cite{AudioSet}  and more zero-shot conditional generation results.
 All results are sampled with 1,000 steps for the best quality.
\noindent\textbf{Unconditional Generation Results}
Firstly, we show more unconditional generation results from Landscape and AIST++ in Figure~\ref{fig:land} and Figure~\ref{fig:aist} respectively. To verify the generative capability of our MM-Diffusion on the open domain, we further train our Coupled U-Net on the largest audio event dataset AudioSet~\cite{AudioSet} with paired videos on the open domain. It contains 2.1 million video clips of 10 seconds, which covers 632 event classes in total. 
The audio in AudioSet is complete and the amount of data is sufficient, but the video quality is not high. 
Therefore, we filter 20k videos of high quality according to the video frame rate and video size. We scale up our Coupled U-Net by enlarging the base channel from 128 to 256, other settings remain unchanged. The visualization results are shown in Figure~\ref{fig:aset}. All corresponding videos in MP4 format are packed in the supplementary video.

\begin{figure*}[t]
  \centering
  \includegraphics[width=\linewidth]{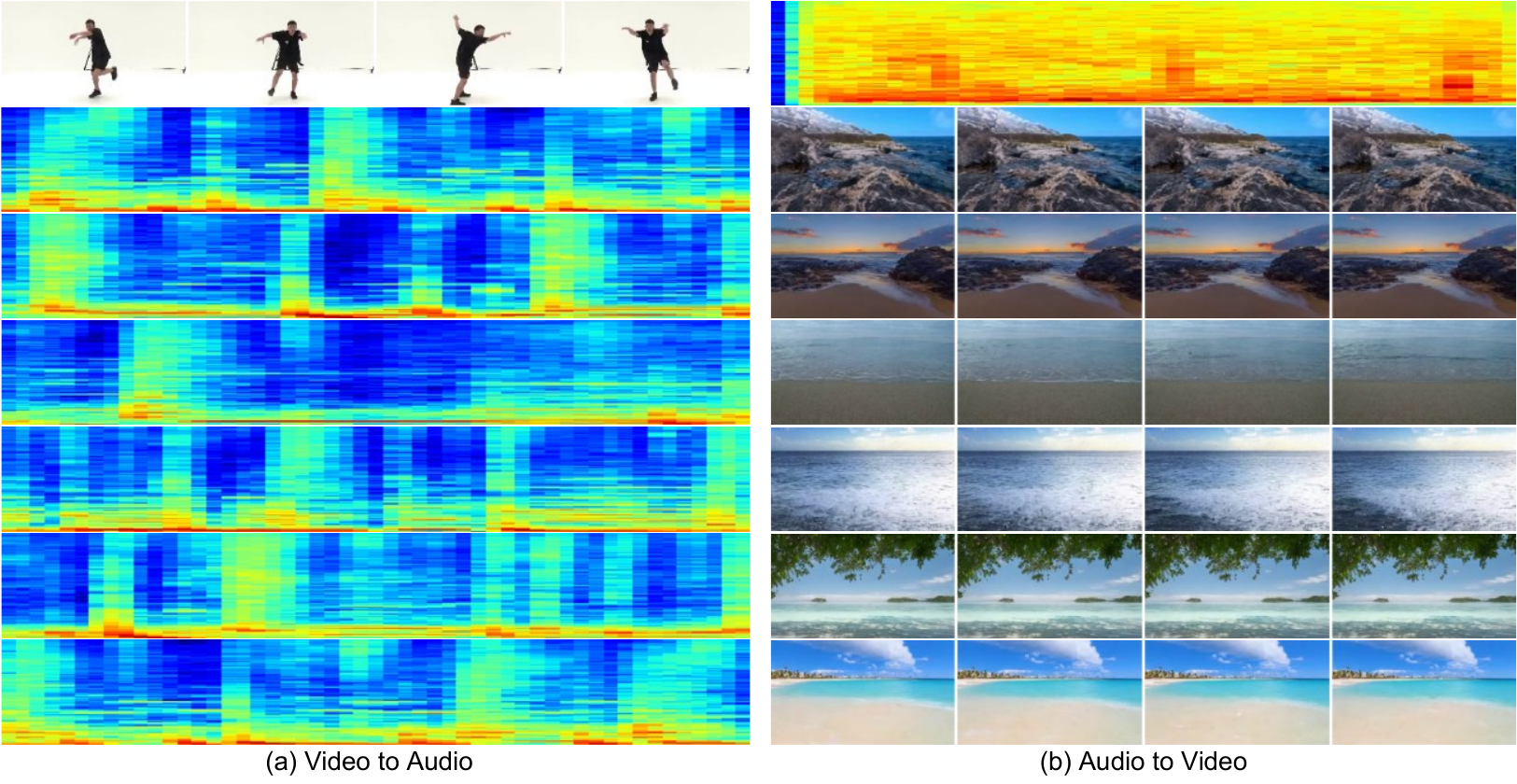}
  \caption{More visual examples of zero-shot conditional generation with our MM-diffusion.}
  \label{fig:avt}
\end{figure*}

\noindent\textbf{Zero-shot Conditional Generation Results}
In this section, we propose video-based audio generation on AIST++. As is shown in Figure~\ref{fig:avt} (a), taking the same video as input, our model can generate different audios corresponding to the dancing beats.
Symmetrically, we next propose audio-based video generation on Landscape. With the results in Figure~\ref{fig:avt} (b), we find that our model can generate diverse video scenes of the sea to the given wave sound. 

\begin{figure*}[t]
  \centering
  \includegraphics[width=\linewidth]{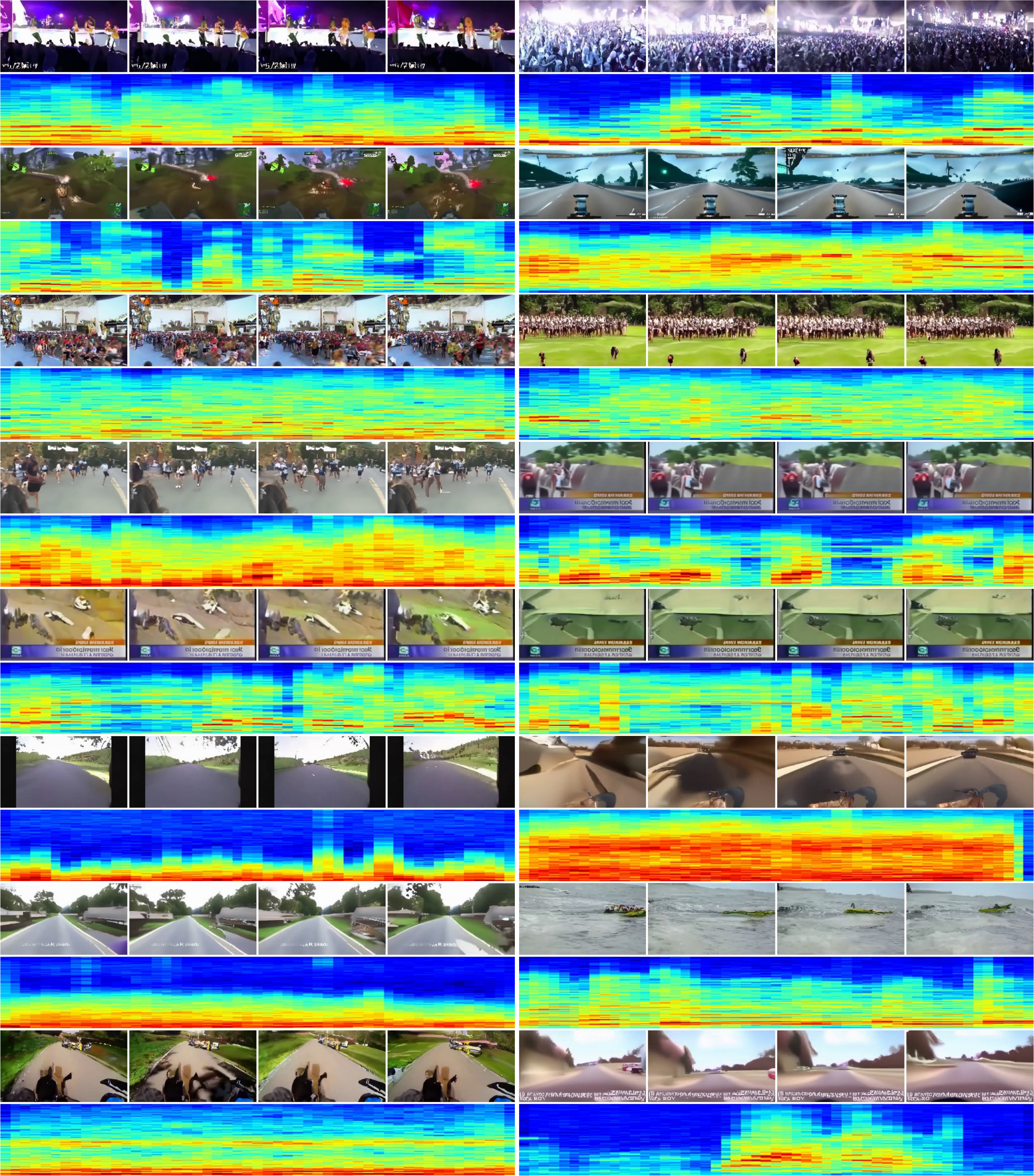}
  \caption{More generation results from open domain~(AudioSet) of our MM-Diffusion.The given cases show the scenes of concert, game streaming, marathon, news playback, surfing and driving in first-person perspective respectively}
  \label{fig:aset}
\end{figure*}

\end{document}